\documentclass[11pt]{article}
\usepackage{times}

\newcommand{\footremember}[2]{%
    \footnote{#2}
    \newcounter{#1}
    \setcounter{#1}{\value{footnote}}%
}
\newcommand{\footrecall}[1]{%
    \footnotemark[\value{#1}]%
} 

\usepackage[top=1in,bottom=1in,left=1in,right=1in]{geometry}

\usepackage{cite}
\usepackage[utf8]{inputenc} 
\usepackage[T1]{fontenc}    
\usepackage{hyperref}       
\usepackage{url}            
\usepackage{booktabs}       
\usepackage{amsfonts}       
\usepackage{nicefrac}       
\usepackage{microtype}      

\title{Understanding racial bias in health using the\\ Medical Expenditure Panel Survey data}

\author{%
  Moninder Singh\footremember{ibm}{IBM Research,
  Yorktown Heights, NY 10598, 
  \texttt{\{moninder,knatesa\}@us.ibm.com}}%
  \and Karthikeyan Natesan Ramamurthy\footrecall{ibm}%
}
\date{}

\begin{document}

\maketitle

\begin{abstract}
Over the years, several studies have demonstrated that there exist significant disparities in health indicators in the United States population across various groups. Healthcare expense is used as a proxy for health in algorithms that drive healthcare systems and this exacerbates the existing bias. In this work, we focus on the presence of racial bias in health indicators in the publicly available, and nationally representative Medical Expenditure Panel Survey (MEPS) data. We show that predictive models for care management trained using this data inherit this bias. Finally, we demonstrate that this inherited bias can be reduced significantly using simple mitigation techniques.
\end{abstract}

\section{Introduction}
Racial and ethnic disparities in access to healthcare in the United States is well-known and documented \cite{NHQDR}. Health disparities are defined to be differences in health outcomes and causes among different groups of people. Health equity is achieved when everyone has the same opportunity to be as healthy as possible.

We have a very good handle on the types of health disparities in the US healthcare system, but the causes for these disparities are complex \cite{nelson2002unequal, nas2018} - such as income, education, socio-economic conditions, neighborhood and community influence, public policy, and societal structure. Achieving health equity also necessitates a complex set of programs and interventions, and several public and private initiatives have tried to address this problem over the past decades.

Obermeyer and Mullainathan \cite{obermeyer2019dissecting} analyze the significant racial bias in an algorithm that drives health decisions for over 70 million people in the United States. They find that black patients with highest health risk have significantly more chronic illnesses than white patients with the same risk score. A key observation they make is that hospitals and insurers treat healthcare expenses as a strong proxy for healthcare needs. This is an imperfect assumption that makes algorithms that predict expenses accurately biased in terms of health. Our work illustrates a similar phenomenon in the publicly available, and nationally representative, Medical Expenditure Panel Survey  (MEPS) dataset \cite{MEPS}.

In this work, we focus on inequity in health indicators in terms of race in the US population. Using the MEPS data, we show that when considering the top decile of high expense white and black patients, the blacks have worse health indicators. We also demonstrate that this carries over to statistical machine learning models for care management that use MEPS data to predict if an individual will incur high expense. Finally, we illustrate how simple bias mitigation methods \cite{Kamiran2012, bellamy2018ai} can be used to make these prediction models fairer, such that blacks and whites in the predicted top decile have lesser health disparity.





\section{Description of MEPS data}
The Medical Expenditure Panel Survey (MEPS) dataset is produced by the US Department of Health and Human Services. It is a collection of surveys of families of individuals, medical providers, and employers across the country. Datasets are available from 1996, and contain two major components: the household component and the insurance component. We use the household component data in this work which contains detailed information on demographic characteristics, health conditions and status, healthcare utilization, access to care, health insurance coverage, income, employment, and charges and sources of payment.

A single panel consists of unique individuals interviewed in five rounds over two full calendar years. Every year, a new cohort is started so that there are two partially overlapping panels being conducted during any calendar year - rounds 1-3 of one panel overlap with rounds 3-5 of the previous panel. In any given dataset, each sample is weighted so that the total weight in each panel sums to the entire US civilian, non-institutionalized population.

\section{Racial bias in the MEPS data}
\label{meps-data-bias}
To explore the presence of racial bias in the MEPS data, we considered 2-year longitudinal data for the cohort initiated in 2015 (panel 20) and surveyed over five rounds during 2015-2016. We restricted the population to individuals who provided data during all five rounds, and indicated their ethnicity/race as non-Hispanic white, or non-Hispanic black. We did a similar analysis of another cohort started in 2014 and covering five rounds over 2014-2015 (panel 19). 

One of the main attributes we studied was total healthcare expenditure. Healthcare expenditure, as a proxy for healthcare utilization, is being increasingly used to identify patients for care/disease management, reducing expenses, as well as evaluating the effects of different policies \cite{fleishman2009,obermeyer2019dissecting,wherry2014,sushmita2015,morid2017}.
From the modeling side, significant amount of work has been done for predicting future healthcare expenses and identifying people likely to incur high medical expenditure, but little has been done in terms of understanding and quantifying the effect of racial disparities.

Since one is typically interested in future healthcare expenditure, we specifically looked at total healthcare expenditure incurred by an individual in the second year (2016). 
We considered the entire population, as well as the top expenditure decile. The 10\% threshold has been commonly used in prior work on building prospective high medical expenditure models \cite{fleishman2009,farley2006}.

In addition to total medical expenditure, we also looked at direct healthcare utilization, as measured by visits to the emergency room (ER) and the number of inpatient nights (IP) during the second year. These outcomes have long been considered as alternatives to total healthcare expenditure, and used to guide various expense and patient management programs as somewhat better proxies for patient health than expenditure \cite{ash2012}. We analyzed these outcome metrics with respect to various individual health metrics in the first year of the panel. Prior work has demonstrated that the count of chronic conditions \cite{fleishman2009} as well as self health assessments and limitations \cite{ash2012, wherry2014, fleishman2009} are significantly associated with future healthcare expenditures: A sicker patient typically has higher resultant medical expenditure/utilization.

In order to obtain a count of chronic conditions for an individual, we looked at the \textit{priority conditions}, a set of conditions, including high blood pressure, diabetes, cancer, stroke, etc., that have been marked as such due to their frequency, expense, as well as importance to policy \cite{PriorityConditions}. We treated four different heart related conditions (coronary heart disease, angina, myocardial infarction, and other unspecified heart disease) as a single condition for the sake of our analysis. We also considered two different self assessed health status measures, one for perceived health status and one for perceived mental health status. Both measures were rated from 1 (excellent) to 5 (poor).

Tables \ref{2016-outcome-metrics-table} and \ref{2015-medical-metrics-table} show the outcome and health indicator metrics for the entire population as well as those in the top decile of healthcare expenditure during the second year.
Clearly, the individuals in the top decile are significantly more expensive, compared to the overall population. More importantly, there is substantial disparity across races, both at the overall level as well as in the high expense group. At the overall level, blacks incur substantially less expenditure, on average, as compared to whites (\$4K versus \$5.9K). This is despite the fact that blacks are typically sicker than whites across almost all metrics. In the top expenditure group, while blacks cost slightly more than whites, they are considerably more sicker than whites, compared to the overall population, as measures both by outcomes (ER visits and IP nights) as well as priority conditions and health status. This echoes the findings of Obermeyer and Mullainathan \cite{obermeyer2019dissecting}, in that blacks at the same risk level are typically much more sicker than whites. Tables \ref{2015-outcome-metrics-table} and \ref{2014-medical-metrics-table} show similar results for the panel 19, 2014-2015 data, although it must be noted that blacks incurred substantially less expense than whites even in the high expense group in that dataset.

More importantly, there is substantial disparate impact across races when it comes to the rate at which blacks and whites incur costs in the top decile. While only 7.1\% of blacks incurred top-decile expense versus 10.7\% of whites in 2015-2016, 6.8\% blacks incurred high expense in 2014-2015 compared to 10.6\% of whites. Thus, not only blacks are sicker than whites at the same risk level, they also incur higher expenses at a lower rate. As we show in Section \ref{before-bias-mitigation}, this bias will likely be inherited by machine learning models designed to identify high expense individuals.

\begin{table}[h]
  \caption{Second year healthcare expenditure and utilization (outcome) metrics for panel 20, 2015-2016 MEPS cohort.}
  \label{2016-outcome-metrics-table}
  \centering
  \begin{tabular}{lllll}    \toprule
     \multicolumn{1}{}{} & \multicolumn{2}{r}{Entire Population}  &   \multicolumn{2}{r}{Top Decile (second}\\
     \multicolumn{1}{}{} & \multicolumn{2}{r}{}  &   \multicolumn{2}{c}{year expenditure)}
     \\    \cmidrule(r){2-5}
    \multicolumn{1}{}{} & \multicolumn{2}{c}{Race}& \multicolumn{2}{c}{Race}
     \\    \cmidrule(r){2-5}  
    Metric     & White     & Black & White & Black  \\     \midrule
        Average expense (both races) & \multicolumn{2}{c}{\$5.6K} &  \multicolumn{2}{c}{\$34.9K}\\
        Top decile expense (both races) & \multicolumn{2}{c}{\$13.6K} &  \multicolumn{2}{c}{}\\
\midrule
        Average expense & \$5.9K& \$4K &  \$34.7K &  \$36.2K\\
    \% of race in top decile & & & 10.7\% & 7.1\% \\
    \midrule
    Average number of ER visits & 0.18 & 0.21 & 0.62  & 0.83    \\
    Average number of IP nights  & 0.33 & 0.45  & 2.61 & 4.91    \\
    \% with ER visits  & 12.9\% &  15.5\% & 40.4\%       & 48\%  \\  
    \% with IP nights  & 6.8\%  &  6.8\% & 44.7\%       & 54.3\%  \\
    \bottomrule
  \end{tabular}
\end{table}

\paragraph{}

\begin{table}[h]
  \caption{First year medical/health-related indicator metrics for panel 20, 2015-2016 MEPS cohort.}
  \label{2015-medical-metrics-table}
  \centering
  \begin{tabular}{lllll}    \toprule
     \multicolumn{1}{}{} & \multicolumn{2}{r}{Entire Population}  &   \multicolumn{2}{r}{Top Decile (Second}\\
     \multicolumn{1}{}{} & \multicolumn{2}{r}{}  &   \multicolumn{2}{c}{year expenditure)}
     \\    \cmidrule(r){2-5}
    \multicolumn{1}{}{} & \multicolumn{2}{c}{Race}& \multicolumn{2}{c}{Race}
     \\    \cmidrule(r){2-5}  
    Metric     & White     & Black & White & Black  \\     \midrule
    Average number of priority conditions & 1.97 & 1.8 &  3.5 & 3.8\\
    Average perceived physical health status & 2.08 & 2.23 & 2.84 & 3.01 \\
    Average perceived mental health status & 1.82 & 1.85 & 2.27 & 2.48\\
    \bottomrule
  \end{tabular}
\end{table}

\begin{table}
  \caption{Second year healthcare expenditure and utilization (outcome) metrics for panel 19, 2014-2015 MEPS cohort.}
  \label{2015-outcome-metrics-table}
  \centering
  \begin{tabular}{lllll}    \toprule
     \multicolumn{1}{}{} & \multicolumn{2}{r}{Entire Population}  &   \multicolumn{2}{r}{Top Decile (second}\\
     \multicolumn{1}{}{} & \multicolumn{2}{r}{}  &   \multicolumn{2}{c}{year expenditure)}
     \\    \cmidrule(r){2-5}
    \multicolumn{1}{}{} & \multicolumn{2}{c}{Race}& \multicolumn{2}{c}{Race}
     \\    \cmidrule(r){2-5}  
    Metric     & White     & Black & White & Black  \\     \midrule
        Average expense (both races) & \multicolumn{2}{c}{\$6.2K} &  \multicolumn{2}{c}{\$40K}\\
        Top decile expense (both races) & \multicolumn{2}{c}{\$15K} &  \multicolumn{2}{c}{}\\
\midrule
        Average expense & \$6.6K& \$4.2K &  \$40.4K &  \$37.5K\\
    \% of race in top decile & & & 10.6\% & 6.8\% \\
    \midrule
    Average number of ER visits & 0.21 & 0.20 & 0.81  & 0.74    \\
    Average number of IP nights  & 0.5 & 0.4  & 4.07 & 4.40    \\
    \% with ER visits  & 14.2\% &  15\% & 42.6\%       & 45.7\%  \\  
    \% with IP nights  & 7.1\%  &  6.8\% & 48.1\%       & 54.4\%  \\
    \bottomrule
  \end{tabular}
\end{table}

\begin{table}[h]
  \caption{First year medical/health-related indicator metrics for panel 19, 2014-2015 MEPS cohort.}
  \label{2014-medical-metrics-table}
  \centering
  \begin{tabular}{lllll}    \toprule
     \multicolumn{1}{}{} & \multicolumn{2}{r}{Entire Population}  &   \multicolumn{2}{r}{Top Decile (Second}\\
     \multicolumn{1}{}{} & \multicolumn{2}{r}{}  &   \multicolumn{2}{c}{year expenditure)}
     \\    \cmidrule(r){2-5}
    \multicolumn{1}{}{} & \multicolumn{2}{c}{Race}& \multicolumn{2}{c}{Race}
     \\    \cmidrule(r){2-5}  
    Metric     & White     & Black & White & Black  \\     \midrule
    Average number of priority conditions & 2.0 & 1.8 &  3.69 & 3.74\\
    Average perceived physical health status & 2.12 & 2.19 & 2.98 & 3.09 \\
    Average perceived mental health status & 1.84 & 1.85 & 2.24 & 2.34\\
    \bottomrule
  \end{tabular}
\end{table}


\section{Predicting individuals that have high expected medical expenditure using the raw MEPS data}
\label{before-bias-mitigation}
We built a logistic regression model to predict second year total medical expenditure of individuals, based on their demographics as well as health-related attributes in the first year. Besides features such as age, gender, and race, we used features on diagnoses received for various priority conditions (high blood pressure, diabetes, heart disease, cancer, etc.) as well as the count of these chronic conditions, physical and mental health assessments, and limitations (such as cognitive or hearing or vision limitation).  We specifically left out certain features such as prior year healthcare expenditure, income, and employment status, that are known to be strong predictors of future healthcare expenditure \cite{wherry2014,morid2017,sushmita2015}. Two  reasons motivated this choice. 
One, from a true care management perspective, it makes sense to relate expected expenditure to factors that can be affected (e.g. chronic diseases) rather than factors that are highly predictive (e.g. prior year expenditure) but are non-actionable. Two, such models, based on diagnosis-related attributes, have been shown to be at least as good in predicting high prospective expenditure individuals as models based on prior expenditures \cite{ash2001}. 

We modeled this problem as a binary classification task - the objective being to predict whether an individual would be in the top decile of second year expenditure. 
The training data for the model consisted of the 2014-2015 Panel 19 data, whereby the model was learned to predict the top decile members of 2015 healthcare expenditure, based on 2014 demographic and health features. The learned model was then applied to the 2015-2016 Panel 20 data to predict the top 2016 expenditure individuals in the cohort based on their 2015 features. The balanced accuracy of the model on the test set was just shy of 73\% (using a threshold obtained from the training data using cross validation). However, to enable a fair comparison with the raw data, the predicted model scores were sorted and only the individuals with scores in the top 10\% were predicted to be high future expense persons.

Table \ref{predicted-outcome-metrics-table} shows some outcome characteristics for individuals predicted to be in the top decile of expenditure in the second year, based on demographics and health conditions in the 
first year. The first point to note, compared to Table \ref{2016-outcome-metrics-table}, is that the racial disparity evidenced in the underlying dataset is picked up by the model as well: only 6.8\% of blacks are predicted to be in the top decile versus 10.6\% of whites. The second point of note is that the corresponding expenses are lower than in the underlying dataset - that is to be expected as features such as prior year expenditure and income which are strong predictors of future expense were explicitly excluded from the feature set. However, the average expense for these individuals is still substantial - around three times the average and higher than the top decile cutoff in the base data.

More importantly, as evidenced from Table \ref{predicted-health-metrics-table-without-bias-mitigation}, these individuals are much more sicker than those in the top decile of expenditure in the underlying dataset. Nevertheless, blacks in this group are sicker, on average, than whites across the measured metrics.
So, while the model as a whole is better suited for care management, as it focuses on individuals who are sicker rather than bigger consumers of healthcare services, it still results in an unfair bias against blacks - not only are they underrepresented in the high risk population, they also have to be sicker than whites to be included.

\begin{table}[h]
  \caption{Second year (2016) healthcare expenditure and utilization (outcome) metrics for predicted high expense individuals in panel 20, 2015-2016 cohort.}
  \label{predicted-outcome-metrics-table}
  \centering
  \begin{tabular}{lll}    \toprule
    \multicolumn{1}{}{} &  \multicolumn{2}{c}{Race} 
    \\    \cmidrule(r){2-3}
    Metric     & White       & Black \\     \midrule
    \% of race predicted to be high-expense & 10.7\% & 6.8\%  \\
    Average expense & \$17.8K & \$16.2K \\
    Average number of ER visits & 0.45  & 0.65       \\
    Average number of IP nights     & 1.32 & 2.58        \\
    \% with ER visits    & 28.6\%       & 39.9\%   \\  
    \% with IP nights    & 20.9\%       & 23.2\%    \\
    \bottomrule
  \end{tabular}
\end{table}

\begin{table}[h]
  \caption{First year (2015) health indicator metrics for predicted high expense individuals in panel 20, 2015-2016 cohort.}
  \label{predicted-health-metrics-table-without-bias-mitigation}
  \centering
  \begin{tabular}{lll}    \toprule
    \multicolumn{1}{}{} & \multicolumn{2}{c}{Race} \\    \cmidrule(r){2-3}
    Metric     & White     & Black  \\     \midrule
    Average number of priority conditions & 4.89 & 5.18 \\
    Average perceived physical health status & 3.55& 3.90 \\
    Average perceived mental health status & 2.53& 2.99\\
    \bottomrule
  \end{tabular}
\end{table}

\section{Predicting high expected medical expenditure from the MEPS data after bias mitigation}
\label{after-bias-mitigation}
To mitigate the racial bias in the MEPS data, and consequently exhibited by the model learned from the data to predict high expenditure individuals, we applied the data preprocessing technique, {\em Reweighing} \cite{Kamiran2012}, to process the training data (2014-2015 Panel 19 MEPS data), to make it more equitable towards blacks. Reweighing works by assigning weights to the tuples in the training data so as to mitigate bias in the data with respect to the sensitive feature, race in our case, without changing the actual labels. The processed Panel 19 data was then used to learn a logistic regression model which was then applied to the Panel 20 data to predict the individuals expected to incur high medical expenditure in the second year (2016), following the same steps as outlined in Section \ref{before-bias-mitigation}. The balanced accuracy of the model trained using the reweighed data a little more than 71 \% on the test set.

The healthcare expenditure and utilization metrics for the predicted, high-expense individuals for the second year are shown in Table \ref{predicted-outcome-metrics-table-with-bias-mitigation}. The corresponding health metrics for those individuals during the first year are similarly shown in Table \ref{predicted-health-metrics-table-with-bias-mitigation}.

The first point to note is the disparity in the racial representation within the predicted high-expenditure group is much less than before: 11\% of blacks are in this group against a slightly lower 9.9\% of whites. This is consistent with the fact that the blacks in this group are still sicker than whites on almost all the metrics (ER visits, IP nights, as well as perceived status), though, importantly, the gap between blacks and whites has reduced on almost every metric compared to before (Tables \ref{predicted-outcome-metrics-table} and \ref{predicted-health-metrics-table-without-bias-mitigation}) - whites are slightly sicker than before while blacks are slightly less sicker than earlier on every metric.

Thus, using a simple bias mitigation technique, we were able to reduce racial bias both in terms of the rate at which the two races were represented in the high risk group and also reduce the gap between them in terms of various measures of sickness.

\begin{table}[h]
  \caption{Second year (2016) healthcare expenditure and utilization (outcome) metrics for predicted high expense individuals in panel 20, 2015-2016 MEPS cohort after bias mitigation.}
  \label{predicted-outcome-metrics-table-with-bias-mitigation}
  \centering
  \begin{tabular}{lll}    \toprule
    \multicolumn{1}{}{} &  \multicolumn{2}{c}{Race} 
    \\    \cmidrule(r){2-3}
    Metric     & White          & Black \\     \midrule
    \% of race predicted to be high-expense  & 9.9\% & 11\% \\
    Average expense &  \$18.4K & \$16.2K\\
    Average number of ER visits     & 0.48  & 0.53    \\
    Average number of IP nights         & 1.41 & 2.33    \\
    \% with ER visits      & 29.9\%       & 33.9\%  \\  
    \% with IP nights      & 21.9\%       & 22.2\%  \\
    \bottomrule
  \end{tabular}
\end{table}

\begin{table}[h]
  \caption{First year (2015) health indicator metrics for predicted high expense individuals in panel 20, 2015-2016 cohort after bias mitigation.}
  \label{predicted-health-metrics-table-with-bias-mitigation}
  \centering
  \begin{tabular}{lll}    \toprule
    \multicolumn{1}{}{} & \multicolumn{2}{c}{Race} \\    \cmidrule(r){2-3}
    Metric      & White     & Black \\     \midrule
    Average number of priority conditions & 4.96 & 4.84\\
    Average perceived physical health status & 3.58& 3.62 \\
    Average perceived mental health status & 2.57& 2.76\\
    \bottomrule
  \end{tabular}
\end{table}

\section{Conclusion and Future Work}
The Medical Expenditure Panel Survey (MEPS) data is a set of publicly available, nationally representative surveys that provides one of the most complete pictures of the expense and utilization of healthcare for the civilian, non-institutionalized population of the United States. One common use of this data is to build predictive models of healthcare expenditure to guide decisions regarding care management, disease management, and cost management. However, none of this prior work has looked at the prevalence of racial bias as it relates to healthcare expenditure and its effect on these models. We show that the bias is also picked up by the models, which results in significant bias against blacks. Namely, blacks are less likely to be predicted as prospective high expenditure patients than white, and hence less likely to be offered care management. Moreover, at the same level of predicted risk, blacks tend to be much more sicker than whites. While this has been noted in other cases \cite{obermeyer2019dissecting}, the fact that MEPS is representative of the entire US healthcare system as a whole, makes this finding even more significant. Furthermore, we show that simple bias mitigation techniques can reduce the bias in models substantially. While it is obviously preferable to use metrics directly related to health conditions and needs, rather than expense, to make decisions about patient and disease management, the fact that various parties remain focused on expense means that understanding the effect of bias and mitigating it in predictive models provides a fairer approach to modeling expenditure.

Nevertheless, this work is still preliminary, and much needs to be done. First, many of the studies involving MEPS data have pooled together data from multiple panels to get larger, more robust data samples. A similar extension of this work is planned.
Second, two components of the MEPS data that have not been used are the medical conditions and event files that provides further detailed information on medical conditions, such as prescriptions, as well as event level details, such as diagnoses received during an ER visit or IP stay. This detailed level data may provide further insight into how pervasive racial bias is across the entire healthcare space. Third, models are also commonly built to predict individuals expected to have high medical utilization such as ER and IP visits, as an alternative to total healthcare expenditure, and used to guide various cost and patient management programs \cite{ash2012}. The presence of racial bias needs to be explored in such models as well.


\medskip

\small
\bibliographystyle{IEEEtran}
\bibliography{paper}

\end{document}